\documentclass[lettersize,journal]{IEEEtran}

\IEEEoverridecommandlockouts

\usepackage{verbatim}
\usepackage{cite}
\usepackage{amsmath,amssymb,amsfonts}
\usepackage{algorithmic}
\usepackage{graphicx}
\usepackage{textcomp}
\usepackage{xcolor}
\def\BibTeX{{\rm B\kern-.05em{\sc i\kern-.025em b}\kern-.08em
    T\kern-.1667em\lower.7ex\hbox{E}\kern-.125emX}}

\usepackage{float}
\usepackage{hyperref}
\usepackage{multirow}

\usepackage{booktabs}

\begin{document}

\title{Proof Number Based Monte-Carlo Tree Search}

\author{\IEEEauthorblockN{Jakub Kowalski\IEEEauthorrefmark{1}, Elliot Doe\IEEEauthorrefmark{2}, Mark H. M. Winands\IEEEauthorrefmark{2},\\ Daniel G\'{o}rski\IEEEauthorrefmark{1}, Dennis J. N. J. Soemers\IEEEauthorrefmark{2}}

\IEEEauthorblockA{\IEEEauthorrefmark{1}\textit{Faculty of Mathematics and Computer Science, University of Wroc{\l}aw}\\\{jakub.kowalski, daniel.gorski\}@cs.uni.wroc.pl}\\
\and
\IEEEauthorblockA{\IEEEauthorrefmark{2}\textit{Department of Advanced Computing Sciences, Maastricht University}\\
e.doe@student.maastrichtuniversity.nl, \{m.winands, dennis.soemers\}@maastrichtuniversity.nl}
}

\maketitle

\begin{abstract}
This paper proposes a new game-search algorithm, PN-MCTS, which combines Monte-Carlo Tree Search (MCTS) and Proof-Number Search (PNS). These two algorithms have been successfully applied for decision making in a range of domains.
We define three areas where the additional knowledge provided by the proof and disproof numbers gathered in MCTS trees might be used: final move selection, solving subtrees, and the UCB1 selection mechanism. We test all possible combinations on different time settings, playing against vanilla UCT on several games: Lines of Action ($7$$\times$$7$ and $8$$\times$$8$ board sizes), MiniShogi, Knightthrough, and Awari. Furthermore, we extend this new algorithm to properly address games with draws, like Awari, by adding an additional layer of PNS on top of the MCTS tree.
The experiments show that PN-MCTS is able to outperform MCTS in all tested game domains, achieving win rates up to  96.2\% for Lines of Action.
\end{abstract}


\section{Introduction}

Monte-Carlo Tree Search (MCTS)~\cite{coulom06,kocsis06b} is a best-first search method guided by the results of Monte-Carlo simulations, well established in game AI. Using the results of previous simulations, the method gradually builds up a game tree in memory and increasingly becomes better at accurately estimating the values of the most promising moves.
The algorithm construction is very open for enhancements, which resulted in classic enhancements \cite{mctssurvey} including MAST \cite{finnsson2008simulation} and RAVE\cite{gelly2011monte}, as well as some newer developments \cite{swiechowski2023monte}, with novel contributions emerging each year \cite{Kowalski2022SplitMoves}. 

MCTS has substantially advanced the state of the art in several deterministic game domains \cite{mctssurvey}, in particular
Go \cite{Silver2017mastering}, but also other board games including  Amazons~\cite{Lorentz08}, Hex
\cite{arneson10}, Lines of Action \cite{winands10},  and the ones of the  General Game Playing (GGP) competition~\cite{bjornsson09}. MCTS has even increased the level of competitive agents in board games with challenging properties such as a larger number of players (e.g., Chinese Checkers \cite{sturtevant08}) or uncertainty (e.g., Kriegspiel \cite{ciancarini10} and Scotland Yard  \cite{nijssen12tciaig}).

In tactical games, where the main line towards the winning
position is typically narrow with many non-progressing alternatives, MCTS may often lead to an erroneous outcome because the nodes' values in the tree do not converge fast enough to their game-theoretic value. To mitigate this effect, MCTS variants have been proposed that integrate concepts of minimax search \cite{winands08b,winands11,LanctotWPS14,baier2015}.

Another promising direction would be the incorporation of  Proof-Number Search (PNS) \cite{allis94} in MCTS. PNS and its variants \cite{vandenHerik2008} have been proposed to prove endgames faster than traditional minimax. PNS variants  have been successfully applied to a large number of domains including
Chess \cite{breuker}, Othello \cite{nagai}, Shogi \cite{nagai}, Lines of Action (LOA) \cite{Winands04}, Go \cite{Kishimoto05b}, Checkers \cite{schaeffer07a}, Connect6 \cite{Wu10}, and the multi-player game Rolit \cite{Saito10}. So far, they have been incorporated in alpha-beta search engines \cite{Winands_2001_Combining}. 
PNS was also combined with pure Monte-Carlo simulations \cite{saito2007monte}.  
All PNS variants share two features: (1) they are algorithms for solving binary goals, such as proving a win or a loss for a game position, and (2) they rely on the concept of proof and disproof numbers.

This paper proposes a new variant, called PN-MCTS, which combines the strengths of MCTS and PNS with each other. The main ideas are to incorporate proof and disproof numbers in the UCB1 mechanism \cite{kocsis06b} of MCTS, and use them for solving subtrees in a similar fashion as MCTS Solver does \cite{winands08b}.
To investigate PN-MCTS performance, we performed game-playing experiments in five two-player, zero-sum board games: Lines of Action (on two different board sizes), \mbox{MiniShogi}, Knightthrough, and Awari, using a variety of different time~settings.

This paper is an extension of work originally presented in \cite{doe2022combining} and contains significant improvements over the methods described there. 
The paper adds a new PN-MCTS enhancement that benefits from solving subtrees and includes extensive tests checking all combinations of proposed enhancements. New experiments cover a wider area of possible settings, and they are computed based on four times more games per test, which notably increases their confidence.
We also propose a novel extension of the originally proposed PN-MCTS algorithm that is able to perform well in games with~draws.

The remainder of the paper is organized as follows. First, MCTS and PNS are discussed in Sections \ref{sec:MCTS} and \ref{sec:PNS}, respectively. Next, we propose a single-layer PN-MCTS in Section \ref{sec:PN-MCTS}. Subsequently, we empirically evaluate the proposed algorithm in the following two sections. First, establishing proper parameter values, and then conducting main experiments testing all combinations of previously defined enhancements.
In Section~\ref{sec:PNMCTS2}, we address the issue of games with draws, proposing an extension of PN-MCTS that can handle them successfully. Finally, Section \ref{sec:Conc} gives conclusions and an outlook on future research.

\section{Monte-Carlo Tree Search}\label{sec:MCTS}

\begin{figure}[hb]
    \centerline{\includegraphics[width= \columnwidth]{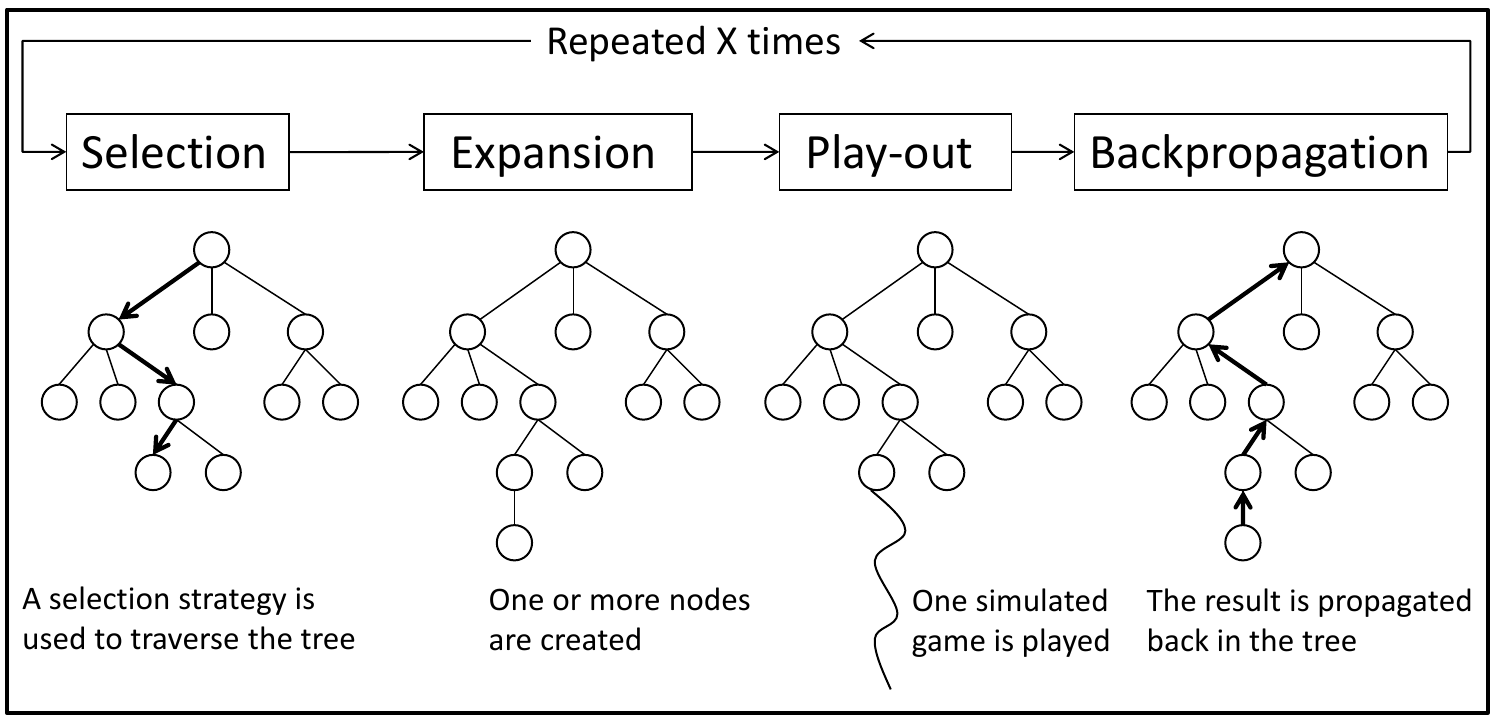}}
    \caption{Outline of Monte-Carlo Tree Search.}
    \label{fig:mcts}
\end{figure}

Monte-Carlo Tree Search (MCTS) \cite{coulom06,kocsis06b} is a best-first search method that does
not require a positional evaluation function. It is based on a
randomized exploration of the search space. Using the results of
previous explorations, the algorithm gradually builds up a game
tree in memory, and increasingly becomes better at accurately
estimating the values of the most promising moves. MCTS consists of four strategic steps, repeated as long as there is
time left \cite{chaslot08}. The steps, outlined in \figurename~\ref{fig:mcts}, are as follows.\\

\textbf{Selection Step.} In the first step, a  child is selected to be searched based on previously gathered information.
The selection step controls the balance between exploitation and exploration. On the
one hand, the task consists of selecting the move that leads
to the best results so far (exploitation). On the other hand, the less promising moves still have to be tried, due to the uncertainty of
the simulations (exploration).

Several \textit{selection strategies} \cite{mctssurvey} have been suggested for MCTS such as BAST, EXP3, UCB1-Tuned, but the most popular one is based on the UCB1 algorithm \cite{auerfinit02}. A standard MCTS implementation with UCB1 as selection strategy is typically called UCT (\textbf{U}pper \textbf{C}onfidence Bounds applied to
\textbf{T}rees) \cite{kocsis06b}. UCB1 works as follows. Let $I$ be the set of nodes immediately reachable from the
current node $p$. The selection strategy selects the child $b$ of
node $p$ that satisfies Formula~(\ref{eq:UCT}):

 \begin{equation}
 \label{eq:UCT}
 \mathit{b\in \mathrm{argmax}_{i \in I} \left(v_i + C \times \sqrt{\frac{\ln{n_p}}{n_i}}\right)},
 \end{equation}

\noindent where $v_i$ is the estimated value of the node $i$, $n_i$ is the
visit count of $i$, and $n_p$ is the visit count of $p$. $C$ is a
parameter constant, which can be tuned experimentally (e.g., $C=\sqrt{2}$).
In the case of a tie, the tie is broken randomly. This process is repeated until a node is reached that has not yet fully been~expanded.\\

\textbf{Expansion Step.} As previously stated, the selection step continues until a node is reached that has not yet expanded all of its children. Among the children that have not been stored in the tree, one is selected uniformly at random. This node $L$ is then added as a  new leaf node and is subsequently~investigated.\\

\textbf{Play-out Step.} From the leaf node the play-out step is performed. Moves are selected in self-play until the end of the game is reached. This step might consist of playing plain random moves or---often better---semi-random moves chosen according to a \textit{simulation strategy}.\\

\textbf{Backpropagation Step.} In the final step, the result \textit{R}
of a play-out $k$ is backpropagated from the leaf node $L$, through the
previously traversed nodes, all the way up to the root. 
Alongside, the visit counter for the visited nodes is increased. 
The result is scored positively $(R_k=+1)$ if the game is won, and negatively
  $(R_k=-1)$ if the game is lost. Draws lead to a result $R_k=0$. A \emph{backpropagation strategy} is applied to the  \textit{value} $v_i$
  of a node $i$. Here, it is computed by taking the average of the results of all simulated games made through this node \cite{coulom06},
   i.e., $v_i=(\sum_{k \in K} R_k ) / n_i$, where $K$ is the set of indices for all play-outs.
\\\\
\indent There are two common ways to select a move to play after the time for MCTS computations run out. One (used in this paper's experiments) is to select a move leading to the root child node with the most visits, and the other is to prioritize the node with the best value (average score) \cite{chaslot08}.

\section{Proof-Number Search}\label{sec:PNS}

 Proof-Number Search (PNS) is a best-first search method especially suited for finding the
game-theoretic value in game trees \cite{allis94}. Its aim is to prove
a particular goal. In the context of this paper, the goal is to prove a forced win for the player to move. A tree can have three values:
\textit{true}, \textit{false}, or \textit{unknown}.  In the case of a
forced win, the tree is \textit{proven}  and its value is true. In
the case of a forced loss or draw, the tree is \textit{disproven}
and its value is false. Otherwise, the value of the tree is unknown.
As long as the value of the root is unknown, the most-promising node
is expanded. Just like MCTS, PNS
does not need a domain-dependent heuristic evaluation function to
determine the most-promising node \cite{allis94}. In PNS,  this
node is usually called the \textit{most-proving} node. PNS
selects the most-proving node using two criteria: (1) the shape of
the search tree (the branching factor of every internal node) and
(2) the values of the leaves. These two criteria enable PNS to
treat game trees with a non-uniform branching factor efficiently.

\begin{figure}[ht]
\centerline{\includegraphics[width=\columnwidth]{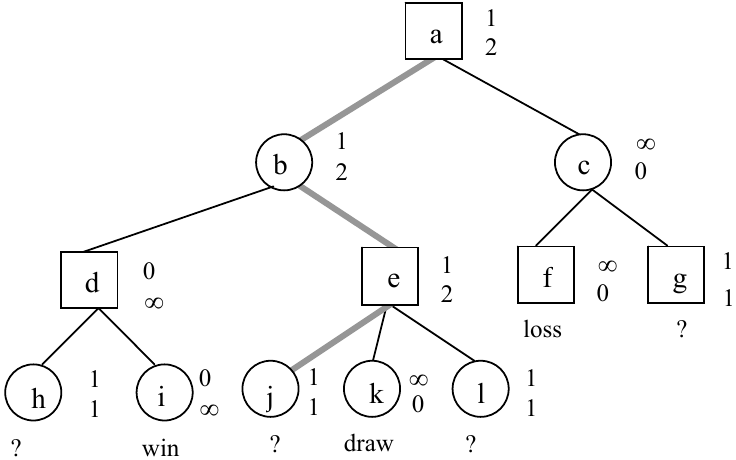} }
\caption{An AND/OR tree with proof and disproof numbers}
 \label{pntree}
\end{figure}

Below we explain PNS based on the AND/OR tree depicted
in Fig. \ref{pntree}, in which a square denotes an OR node, and a
circle denotes an AND node. The numbers to the right of a node
denote the proof number (upper) and disproof
number  (lower). A \textit{proof number}
(\emph{pn}) represents the minimum number of leaf nodes, which have
to be proven in order to prove the node. Analogously, a
\textit{disproof number} (\emph{dpn}) represents the minimum number
of leaf nodes that have to be disproven in order to disprove the
node. Because the goal of the search is to prove a forced win, winning
nodes are regarded as proven. Therefore, they have $pn=0$ and
$dpn=\infty$ (e.g., node \textit{i}). Lost or drawn
nodes are regarded as disproven (e.g., nodes \textit{f} and
\textit{k}). They have $pn =\infty$ and $dpn=0$.
Unknown leaf nodes have  $pn=1$ and $dpn=1$ (e.g.,
nodes \textit{g}, \textit{h}, \textit{j}, and \textit{l}).  The \emph{pn} of an
internal OR node is equal to the minimum of its children's proof
numbers, since to prove an OR node it suffices to prove one child.
The \emph{dpn} of an internal OR node is equal to the sum of
its children's disproof numbers, since to disprove an OR node all
the children have to be disproven. The  \emph{pn} of an internal AND node is equal to the sum of its children's
proof numbers, since to prove an AND node all the children have to
be proven. The \emph{dpn} of an AND node is equal to the
minimum of its children's disproof numbers, since to disprove an AND
node it suffices to disprove one child. 

The procedure of selecting the most-proving node to expand next is as
follows. The algorithm starts at the root. Then, at each OR node the child with
the smallest \emph{pn} is selected as successor, and at each AND
node the child with the smallest  \emph{dpn} is selected as
successor. Finally, when a leaf node is reached, it is expanded
(which makes the leaf node an internal node) and the newborn
children are evaluated. This is called
\textit{immediate evaluation}. The selection of the most-proving
node (\textit{j}) in Fig. \ref{pntree} is given by the bold path.

\section{PN-MCTS Algorithm}\label{sec:PN-MCTS}
To determine a feasible approach to incorporating (dis)proof numbers into MCTS, it is first important to consider what information the proof and disproof numbers bring. As explained in the previous section about PNS, a (dis)proof number provides a lower bound for the number of nodes that still have to be (dis)proven to prove the current node. 

In PNS, these lower bounds determine which leaf node would be investigated further. In MCTS, the selection step has a similar function. In the default MCTS implementation, this would use the UCB1 Formula (\ref{eq:UCT}). Thus, a natural way to combine MCTS and PNS would be to combine these two ways of selecting promising leaf nodes and modify the basic UCB1 formula with knowledge gained from PNS. 

A direct consequence of tracking this lower bound is knowledge about the proven/disproven subtrees. Although for any game of proper size we should not expect the game root to be proven, it may be possible that near the endgame, the current state became provable. Also, many subtrees can be (dis)proven within the MCTS tree during a search, so there is no need to revisit them again during the expansion phase.

Therefore, we introduce three PNS-based types of MCTS enhancements, regarding \emph{Final move selection}, \emph{Solving subtrees}, and \emph{UCB1 formula}. They can be applied independently, in the form of boolean flags (each enhancement turned on/off).

\subsection{Final Move Selection}

This is a simple yet effective enhancement. Due to the proof numbers stored in nodes, we know which of the subtrees rooted in these nodes are proven to win for our player. Thus, if during the final move selection there is a move leading to a proven child, we should always select it regardless of the number of visits / average score in this node.

We may be tempted to use disproof numbers in a similar way, forbidding the selection of such nodes if any alternative exists, but please note that if the game has more than two outcomes, all proven outcomes other than a win will count as disproven. E.g., we will not distinguish a node that is a loss from one that leads to a draw. We present a solution to this issue in Section~\ref{sec:PNMCTS2}.

\subsection{Solving Subtrees}

Proof and disproof numbers can be seen as an extension of a solver \cite{winands08b,cazenave2011ScoreBoundedMCTS}, a successful MCTS enhancement that is obligatory in most applications. MCTS-Solver backpropagates information from leaf nodes being terminal states and marks subtrees as won, lost, or unresolved. PN-MCTS computes a superset of this information, so it can be used in the same way---to save the computational effort and skip revisiting already (dis)proven subtrees.

In practical applications of MCTS-Solver, usually an additional parameter $T$ is introduced to deal with bias favoring narrow paths to win during the selection step when some children of a node are solved. Thus, a node is omitted during the UCT selection only if the number of visits in this node exceeds the threshold $T$ \cite{winands08b}. In the case of all our experiments, if the solver enhancement is on, $T$ is equal to~5.

\subsection{UCT-PN}

The third enhancement proposed in this paper is a modification of basic UCB1. By adjusting UCB1 to also use proof and disproof numbers, the information from the (dis)proof number can influence the decision making process, with no need to change anything else about MCTS.

The final consideration then is how to use (dis)proof numbers in UCB1. 
The magnitudes of differences amongst (dis)proof numbers technically do not have much meaning. For example, a node with a proof number of 100 is not necessarily ten times worse than a node with a proof number of 10. The node with the proof number of 100 may just have been investigated more often already. The fact that the magnitudes of differences between (dis)proof numbers do not have much meaning makes it difficult to directly use them in the UCB1~formula.

Instead of using the proof or disproof numbers directly in the formula, this paper proposes that the (dis)proof numbers are used to determine a ranking amongst all the nodes. The ranking is similar to the one of PNS as explained in Section \ref{sec:PNS}. At an OR node the child node with the lowest proof number would get the best ranking because that is the node that would be selected in regular PNS. 
For example, if there are 30 child nodes, the one that would be picked according to PNS gets a rank of 1. 
At an AND-node, the node with the lowest disproof number is picked. The next best ranking node would be the node PNS would pick if the original best ranking node was not an option (so the second lowest (dis)proof number would get a ranking of 2, whereas the worst option would get 30 for this example). Ties are awarded the same rank. Finally, this rank can then be normalized to be in the range of [0, 1].

Normalization allows the resulting value to be in a range that is similar to the values that might come out of the exploitation or exploration parts of the UCB1 formula, in a manner that is also automatically adaptive to differences in the number of legal actions between different states.
Our adjusted UCB1 formula, referred to as the UCT-PN formula looks as follows:

\begin{scriptsize}
\begin{equation} \label{UCT-PN_formula}
 \mathit{b\in \mathrm{argmax}_{i \in I} \left(v_i + C \sqrt{\frac{\ln{n_p}}{n_i}} + C_{pn} \left( 1 - \frac{pnRank_{i}}{\mathrm{max}_{j \in I} (pnRank_{j})} \right) \right)}.
\end{equation}
\end{scriptsize}

\noindent 

To normalize, the rank of a specific node $pnRank$ is divided by the largest rank value of any of the children. The lowest rank means best node according to PNS, so $max_{j \in I} (pnRank_{j})$ is the highest (and thus worst) rank of any child node. 
To control the influence of the addition, the PN-Parameter $C_{pn}$ is added.
The rest of the variables are the same as in the regular UCB1 Formula~(\ref{eq:UCT}). 
This paper uses the term PN-MCTS to refer to any MCTS variant that uses the UCT-PN formula instead of the base UCB1 formula for its selection step.

\section{Environment}

Here, we briefly introduce the domains used to test our algorithm.
We outline the general game-playing system Ludii, in which all algorithms and games were implemented, and explain the choice for this system.
Then, we present the rules of the five games used as a test domain to compare our PN-MCTS with the basic MCTS.

\subsection{Ludii General Game System}\label{ludii}

The Ludii General Game System \cite{Piette2020Ludii} is a general game-playing framework, which provides an environment for developers to test their implementation of general game-playing agents. Its advantage over the other similar systems (\cite{Genesereth2005General,Kowalski2019RegularBoardgames}) comes from including over 1,000 games described in its game description language, and implementations of various standard algorithms and enhancements (such as several variants of MCTS).
It has a single, unified API for the development of intelligent agents, based on a forward model (with functions to generate lists of legal actions, generate successor states, and so on) and standardized state and action representations.
Ludii has been demonstrated \cite{Piette2020Ludii} to process games faster than the previous state-of-the-art general game-playing framework based on Stanford's Game Description Language \cite{Love_2008_GDL}, which is important for the playing strength of game-search algorithms such as MCTS.

\subsection{Game Domains}\label{domains}

Two-player adversarial games are well suited to PNS as it structures its knowledge as AND/OR-trees. The list of games that fulfill this condition is still rather large. 
To narrow the list down even more, only domains in which both MCTS and PNS have shown to be effective are considered.
If either MCTS or PNS does not perform well in a domain, the combination of the two will probably not be very effective.
From the remaining list of games that fits the requirements and desirable qualities, four games are chosen: Lines of Action, Awari, MiniShogi, and Knightthrough. 
Each of the games is briefly described~below. 

\subsubsection{Lines of Action}

The rules of Lines of Action (LOA) are as follows \cite{sackson69}. It is played on an
8$\times$8 board by two sides, Black and White. Each side has twelve
pieces at its disposal. The black pieces are placed along the top and bottom rows of the board, while the white pieces are placed in
the left- and right-most files of the board. The players alternately move a piece, starting with Black. A piece moves in a straight line, exactly as many squares as there are pieces of either color anywhere along the line of movement. A player may jump over its own pieces, but not the
opponent's, although opposing pieces are captured by landing on them.
The goal of the players is to be the first to create a configuration on the board in which all their pieces are connected in a single group (with each piece being directly adjacent to at least one of the other pieces). Adjacencies may be orthogonal or diagonal.

There are two main reasons why LOA was chosen as the main test domain. Both MCTS and PNS have been extensively tested on LOA \cite{vandenHerik2008,winands10}, and the game board has an adjustable size. The default board is 8$\times$8, but  smaller sizes such as 7$\times$7 can also be used. The advantage of the smaller board sizes is that the game reaches endgame states much quicker. PNS works best in endgame scenarios where game states can be proven or disproven in fewer steps. Thus, by testing on various board sizes, the experiments can test whether PN-MCTS has a better performance when endgame states require fewer steps to be reached.

\subsubsection{Awari}
Awari is a Mancala or sowing game \cite{awari}. It is is played on a 2$\times$6 board and with counters. The goal of the game is to capture as many counters as possible. To capture counters, a player must end their sow in the opponent's row and in a hole with 2 or 3 counters (including the piece used to sow). Sowing is a process where a player takes all the counters from a hole in their row and deposits them one by one into adjacent holes until none are left. In Awari, sowing goes counter-clockwise. The game is over once none of the holes contain more than 1 counter. The player who captured most counters wins, or in case both players have an equal amount, the game ends in a draw.  Awari is a suitable test domain for PN-MCTS as Mancala variants have been used as testbed for PNS \cite{allis94} and MCTS \cite{LanctotWPS14} in the past.

\subsubsection{MiniShogi}
MiniShogi is a variant of the old Japanese game called Shogi and was invented around 1970 by Shigenobu Kusumoto. The game has various pieces each of which have their own rules for movement. The goal is to capture the opponent's King with these pieces. Pieces can be promoted by moving them towards the opponent's side of the board. Opponent pieces can be captured by moving a piece onto an enemy piece. Instead of moving a piece, players may also spend their turn by placing a piece they have previously captured as a new piece of their own on the board. MiniShogi differs from regular Shogi in the following ways: it is played on a 5$\times$5 board, it has fewer pieces than the original, and features a smaller promotion area. Shogi endgames have  been one of the main test domains of PNS \cite{KishimotoW0S12}, whereas MCTS has become the dominating search technique for this game \cite{SilHub18General}.

\subsubsection{Knightthrough}
The game of Knightthrough is a variant of Breakthrough \cite{handscomb01}. It is played on an 8$\times$8 board. Each player has 16 pieces in the first two rows of their side of the board (opposing sides). Every piece moves like a knight in chess. This means each piece may move 1 square in one non-diagonal direction and then 2 squares in a perpendicular direction. Pieces may be captured by landing on them. Knights can jump over other pieces (both friendly and opponent). The goal of the game is to reach the opponent's edge of the board (the row furthest from the player) with one of their knights. A player can also win by capturing all opposing pieces. Knightthrough has been used to test MCTS in a general-game-playing context \cite{SironiLW20}. Its original variant Breakthrough has served as a test bed for PNS variants \cite{saffidineJC11}.

\section{Experiments}\label{sec:PremExp}

This section outlines the experiments that have been conducted for the initial parameter tuning and measurement of computational efficiency difference between the compared algorithms. 

The variant of PN-MCTS with final move selection and UCT-PN formula (solver enhancement is disabled) is tested against a basic UCT implementation from the Ludii system.\footnote{\url{https://github.com/Ludeme/LudiiExampleAI}} The PN-MCTS extension has been built on the same code base. For both agents the MCTS $C$ parameter is set to $\sqrt{2}$.

For every result, if the error margins are presented, they mean a 95\% confidence interval. When calculating winrates, we count draws as 0.5 wins.  Also, player positions are swapped in all tests so that the agents play both sides equally~often.
 
All of the experiments described in this paper have been run on an Intel(R) Core(TM) i7I7-11700K CPU with 2$\times$ 16GB RAM. 
The repository containing the source code of the algorithms presented in this work and exact encodings of the games used is available on GitHub.\footnote{\url{https://github.com/acatai/pn-mcts}}

\subsection{PN-Parameter}
First, we want to tune the $C_{pn}$ parameter. Values of $C_{pn} \in \{0.0, 0.1, 0.5, 1.0, 2.0, 5.0,  10^6\}$ are tested, each with 1000 games against the base MCTS. One of the tested $C_{pn}$ configurations is set to an arbitrarily high number $10^6$. Then, the PN-MCTS performs semantically the same as a regular PNS (with the additional overhead of Monte-Carlo simulations), except that ties in proof and disproof numbers are broken by UCB1 instead of randomly. The experiments are executed with 1 second per turn. This experiment is conducted in the game of LOA on the board sizes 7$\times$7 and 8$\times$8. By playing on two board sizes, results can be compared to determine if larger search spaces, longer games, and later endgame situations have different trends in which parameters are best for the~\mbox{PN-MCTS}.

\begin{figure}[t]
    \centering
    \includegraphics[width= \columnwidth]{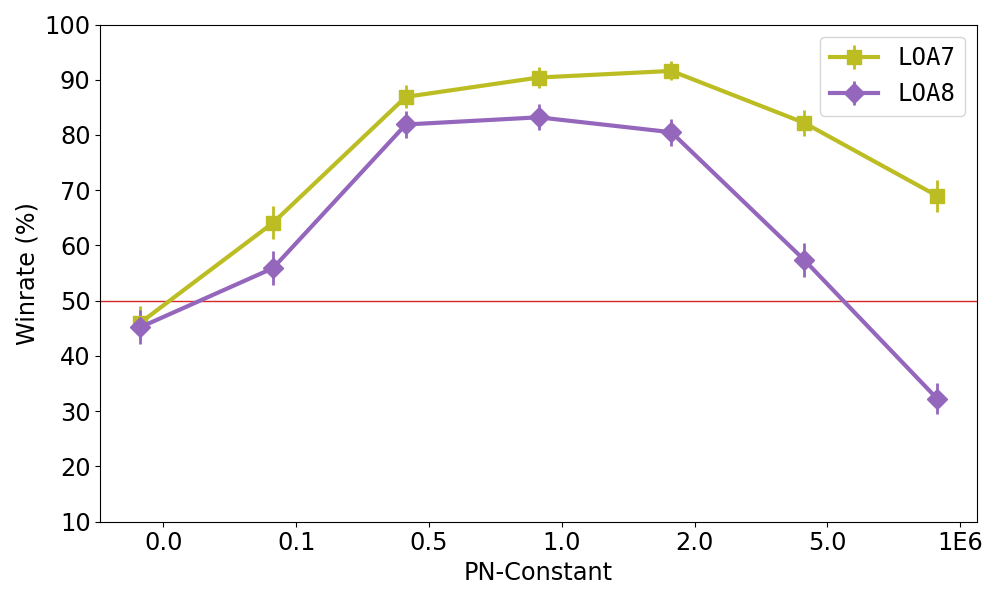}
    \caption{Tuning the $C_{pn}$ parameter. PN-MCTS against MCTS for 7$\times$7 and 8$\times$8 LOA (1000 games, 1s per turn).}
    \label{fig:cpntuning}
\end{figure}

Figure \ref{fig:cpntuning} displays the win rates of PN-MCTS with varying $C_{pn}$ against MCTS. The trend in the win rate of PN-MCTS is fairly similar for both board sizes. 
If $C_{pn}$ nears 0, such as when it is 0.1, the PN-MCTS will behave more like the basic MCTS with final move selection (so it makes up for decreased efficiency).
Another low point for both board sizes is when $C_{pn}$ is $10^6$. When $C_{pn}$ is that high, the agent starts behaving similarly to PNS. From the results, it seems that a basic PNS still performs better than basic MCTS on the smaller 7$\times$7 board, but not on the regular 8$\times$8 board. This is expected as PNS is most effective in the endgame.  
Due to the endgame being reached sooner on the 7$\times$7 board, endgame states---in which PNS and MCTS-PNS with high $C_{pn}$ are more effective---constitute a greater proportion of the game tree.

As for the highest win rates, on a 7$\times$7 board, a $C_{pn}$ of 2.0 performs best, recording a win rate of 91.2\% over basic MCTS. On the 8$\times$8 board, a $C_{pn}$ of 1.0 appears to be the optimal one, with a win rate of 83.2\%. 

$C_{pn}$ represents the impact that the (dis)proof numbers have on the basic MCTS. So on the smaller board size, where PNS performs better than MCTS, a larger influence of the proof and disproof number performs better. This is only true to some degree, as when $C_{pn}$ reaches 5.0, the win rate drops as it starts to converge toward the score of a pure PNS. 

To give an idea how a pure PNS would behave on its own without any CPU overhead, we conducted an experiment with $C_{pn}=10^6$ and a fixed number of 1000 simulations per move in LOA 8$\times$8. Here PN-MCTS won only 9 out 100 games against the regular MCTS. For the same number of simulations, but for $C_{pn}=1$, PN-MCTS won 99 out 100 games. These results validate that PNS on its own is weaker than MCTS, but when combined together in PN-MCTS the algorithm may outperform MCTS.

\subsection{Overhead Estimation}

This experiment investigates the cost of obtaining, and continuously updating, the proof and disproof numbers. PN-MCTS is constructed as an augmentation of the basic MCTS implementation provided by the Ludii system to ensure that any difference in performance is solely due to the implementation of the proposed enhancement.
The (dis)proof numbers have to be recalculated on backpropagation; thus, in general, the larger the tree, the higher the backpropagation has to go and the higher the processing cost should be. 

To obtain the cost specifically, we count and compare the number of simulations performed by PN-MCTS and the original MCTS.
For each of the five tested games, we played 1000 matches with 1-second turn limit, summing the number of simulations performed during the first half of the match. We applied this methodology to remove noise created by the end-game search, usually consisting of a huge number of iterations with zero-depth playouts.
We take the average number of simulations per match per algorithm and obtain a ratio dividing the number of PN-MCTS simulations by MCTS simulations. Thus, the smaller the number, the slower our implementation of PN-MCTS in this game compared to the basic MCTS. The results of the overhead experiment are presented in Table \ref{tab:simsPerSecond}.

The table reveals that PN-MCTS has a relatively mild overhead for three out of five tested games, where ratios are no lower than $0.9$.
For Knighthrough and MiniShogi the drop is more significant. For these games, efficiency is a disadvantage that would need to be overcome by the quality of the PN~enhancement.

There is one difference in our PN-MCTS implementation compared to basic MCTS that can have an influence here. PN-MCTS fully expands all children of nodes at once---allowing for the computation of (dis)proof numbers and rankings for all children---whereas the standard MCTS only expands one child per simulation.
This increases memory usage for PN-MCTS and hinders its performance for games with large branching factor and a relatively high baseline number of simulations per second (e.g., Knightthrough). 

The extreme case encountered during our research was related to Gomoku\cite{gomoku}.
It is a connection game, played on a 15$\times$15 board, with black and white stones. The goal is to make a row of exactly five stones of the same color. 
As PNS has been applied in Gomoku before \cite{allis96}, and MCTS variants have also been developed for this game \cite{TangZSL16}, we included it in our initial experiments. 
However, as the game has quite a large branching factor (initially 225) and a simple (thus fast) move application, we obtained an overhead ratio of 0.054, so our implementation was shown to be nearly 20 times slower. 
For this reason, we do not include the remaining results concerning~Gomoku.

It is important to note that this particular behavior is not required by the PN-MCTS enhancement itself. We plan to improve our implementation in this regard in future work.

\begin{table}[H]
\begin{center}
    \caption{PN-MCTS overhead over MCTS, based on the average number of simulations performed per game (1000 games, 1s per turn)}
    \label{tab:simsPerSecond}
    \begin{tabular}{@{}lrrr@{}}
    \toprule
    Game & PN-MCTS sims & MCTS sims & PN-MCTS/MCTS ratio \\
    \midrule
    LOA 7$\times$7 & 8401.306  & 8811.572  & 0.953\\
    LOA 8$\times$8 & 16804.794 & 18549.838 & 0.901\\
    MiniShogi      & 573.641   & 774.442   & 0.741 \\
    Knightthrough  & 15376.921 & 19919.245 & 0.772 \\
    Awari          & 30395.64  & 31722.801 & 0.959 \\
    \bottomrule
    \end{tabular}
\end{center}
\end{table}

\section{PN-MCTS Experiments}

In this section, we present two groups of experiments. The goal of the first one is to measure the influence of each enhancement of PN-MCTS on the overall result. Thus, we test all combinations of these enhancements using a time limit of 1 second per turn. The second experiment aims to show the trends of the algorithm given different time limits. For this test, we chose the most successful variants of PN-MCTS based on the previous experiment.

We played 1000 games for each test (with 500 games per side), and $C_{pn}=1$.
To encode which PN-MCTS enhancement is turned on, we use the following notation. Final move selection is encoded by \verb|F|, Solver by \verb|S|, and UCT-PN formula by \verb|U|. If an enhancement is off, the \verb|x| symbol is used instead. Thus, \verb|xSx| encodes a variant where only PN-solver is used, while \verb|FSU| is a variant with all proposed enhancements enabled.

\subsection{PN-MCTS Variants}

The results of the experiment are presented in Table~\ref{tab:variants}. To make the comparison meaningful, in the \verb|xxx| variant, none of the enhancements are used, but all data required to use them is computed. Thus, this is the baseline, in which the winrates are strictly correlated with the results of the overhead experiment.

There are two main observations. First is that the behavior of the PN-based enhancements is rather game dependent, and as it may be quite beneficial for some games, it may also be quite harmful for others. 
Second is that, indeed, proposed enhancements are mostly improvements, and turning them on increases the winrates. It is not always the case, but the \verb|FSU| variant with all enhancements is the best combination in 3 out of 5 tested domains and second-best in one more.

\begin{table*}[t]
  \centering
\caption{Comparison of PN-MCTS variants. Given winrates are against a basic MCTS. (1000 games, $C_{pn}=1$, 1s per turn). Enhancements encoding -- Final move selection: \texttt{F}, Solver: \texttt{S}, UCT-PN: \texttt{U}, no enhancement: \texttt{x}.}
\label{tab:variants}
\begin{tabular}{l|rrrrrrrr}
\toprule
 \multirow{2}{*}{Game domain}& \multicolumn{8}{c}{PN-MCTS variant} \\
                & \verb|xxx|  & \verb|Fxx|   & \verb|xSx|  & \verb|xxU|  & \verb|FSx|  & \verb|FxU|  & \verb|xSU|  & \verb|FSU| \\ \hline 
LOA $7\times 7$ &$49.8\pm3.10$& $46.0\pm3.09$&$61.0\pm3.02$&$87.8\pm2.02$&$65.6\pm2.95$&$90.4\pm1.83$&$91.1\pm1.76$ & $\textbf{93.2}\pm1.56$\\ 
LOA $8\times 8$ &$44.6\pm3.08$& $45.2\pm3.09$&$63.1\pm2.99$&$86.2\pm2.13$&$66.4\pm2.93$&$83.2\pm2.32$&$\textbf{92.9}\pm1.59$ & $90.8\pm1.79$ \\
MiniShogi       &$31.3\pm2.88$& $48.2\pm3.10$&$40.9\pm3.05$&$55.5\pm3.08$&$53.2\pm3.09$&$64.5\pm2.97$&$60.3\pm3.03$ & $\textbf{65.8}\pm2.94$ \\
Knightthrough   &$43.1\pm3.07$& $52.3\pm3.10$&$55.9\pm3.08$&$51.6\pm3.10$&$59.2\pm3.05$&$56.4\pm3.08$&$63.5\pm2.99$ & $\textbf{66.8}\pm2.92$  \\
Awari           &$34.2\pm2.74$& $34.4\pm2.77$&$35.2\pm2.86$&$48.2\pm2.69$&$38.0\pm2.94$&$\textbf{49.8}\pm2.72$ & $40.0\pm3.01$ & $40.6\pm3.01$ \\
\bottomrule
\end{tabular}
\end{table*}

\subsubsection{Final Move Selection}

Although the benefits of this enhancement are apparent, and it has always been a crucial part of MCTS-Solver \cite{winands08b}, its impact was never measured alone, apart from the selection part of the solver. Our experiments measure this impact, as we can directly compare combinations of enhancements with, and without final move selection: \verb|xxx| and \verb|Fxx|, \verb|xSx| and \verb|FSx|, \verb|xxU| and \verb|FxU|, finally \verb|xSU| and \verb|FSU|.

There are a few interesting observations here. The first is that there are three cases where this enhancement lowers the winrate, all of which are for LOA. While these reductions are by statistically significant margins according to one-sided tests with 95\% confidence levels, the effect sizes are relatively small (the worst case is a drop of 3.8 percentage points, from 49.8\% to 46.0\%). The effect is also not necessarily consistent, in the sense that there are also other LOA cases with an effect in the other direction.

In all other cases, selecting a proven move gives an additional few percent points. Usually, the simpler the algorithm (more enhancements are off), the larger the impact of the enhancement. In some cases, for MiniShogi and Knightthrough, the benefits may reach 10\%.
Lastly, the enhancement does not benefit so much drawish games as Awari, as it does not influence the decision of when to draw instead of possibly losing (improvements up to 3\%). We come back to this topic in the next section.

\subsubsection{Solving Subtrees}

This enhancement is usually beneficial, and quite impactful, providing improvements ranging from about 3 percent points with all other enhancements, to nearly 20 percentage points for LOA and simpler variants of the algorithm, e.g., \verb|Fxx| to \verb|FSx|.
For MiniShogi and Knighthrough, it is also a clear improvement, reaching up to 10 additional percent points winrate.

However, for Awari, solver tends to be harmful. The highest negative impact observed, over $-9\%$ is for the most complex variant, between \verb|FxU| to \verb|FSU|. Again, the reason here is the high frequency of draws in the game. Most likely, the impact of excluding of some child nodes from contributing to the UCB1 selection formula.

\subsubsection{UCT-PN}

For all the tested games and relevant variant pairs, extending the UCB1 formula with UCT-PN proves to be beneficial. All overall best results for each game contain this enhancement. 
In 14 out of 20 comparable variants, winrates increased by more than 10\%. The biggest observed improvement, 44.4\%, is for LOA $7\times7$, from \verb|Fxx| to \verb|FxU|.  

On the one hand, such results should be expected, as the test set contains games where PNS has proven to be effective. On the other hand, it is not necessarily true that combining two valid approaches will result in an even better one. Also, the ranking system UCT-PN introduces, requires sorting nodes, which is the potentially costly part of the proposed extension.

Overall, PN-MCTS wins in 4 out of 5 tested games against the standard UCT-MCTS, with a clear over 90\% winrate for LOA, and a tie for Awari.

\subsection{Time Trends}
\begin{figure*}[t]
    \centering
    \includegraphics[width= \textwidth]{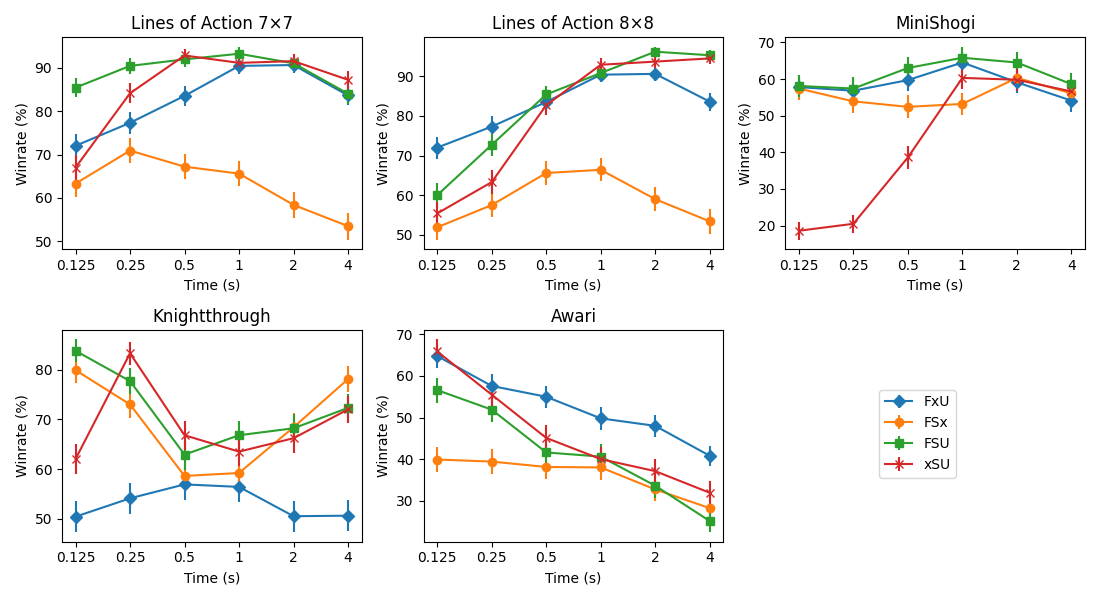}
    \caption{Comparison of the best PN-MCTS variants in different time settings, playing against a basic MCTS (1000 games, $C_{pn}=1$).}
    \label{fig:timetrends}
\end{figure*}

To assess the usefulness of the proposed MCTS variant it is necessary to measure the effect of time and estimate the trends. We have selected all variants with at least two enhancements applied (\verb|FxU|, \verb|FSx|, \verb|FSU|, \verb|xSU|), and run them for different time limits: 0.125, 0.25, 0.5, 1, 2 and 4 seconds per move. The results of this experiment are presented in Figure~\ref{fig:timetrends}.

The more computation time both algorithms have, the greater impact of the difference in efficiency, and relatively pure MCTS makes many more simulations. However, the PN component of the PN-MCTS has the most impact in endgame positions, so with more time, there is a higher chance of reaching them sooner. The trends depend on the balance of each of these factors for each of the tested domains.

For LOA, independently of the board size, there is generally an improving tendency for variants with the UCT-PN enhancement, and worsening for \verb|FSx|. 
The peak winrate with 96.2\% is reached by \verb|FSU| for the larger LOA board and the 2 seconds time limit.
For MiniShogi variants with final move selection perform roughly stable for all times, while \verb|xSU| has poor results for low time limits, and becomes comparable to other variants with $\geq 1$s.
The trends for Knightthrough are harder to generalize, with \verb|FSx| and \verb|FSU| behaving in U-like shape, and \verb|FxU| being clearly worse, staying below 60\% for all times.
The general behavior for Awari is that with more time, the results of PN-MCTS are worse for all variants tested.

\section{Double-Layer PN-MCTS}\label{sec:PNMCTS2}

As already mentioned, PN-MCTS in the form presented above does not properly handle games with draws: proven nodes are won nodes, but disproven nodes can be drawn or lost. However, we have a solution for this issue.

Instead of remembering just a proof and disproof number for each MCTS tree node, we remember two additional values: second-layer proof number and second-layer disproof number. These data are processed exactly as the first-layer (dis)proof numbers described before, with one important difference: now the semantic of our proof is \emph{``not lost''}.  The resulting game-theoretic values of each (dis)proof number are presented in Table~\ref{tab:2layersemantic}.

We found our concept somewhat similar to the one in \cite{saffidine2012multiple}. However, as our PNS application is an online search, rather than an offline solver, and the proposed solution is determined by the choice of the UCT-PN formula, some of the optimization from their publication cannot be used. 

\begin{table}[h!!]
\begin{center}
    \caption{Game-theoretic values for double-layer PN-MCTS}
    \label{tab:2layersemantic}
    \begin{tabular}{c|c|c}
    \toprule
     & Proven nodes & Disproven nodes  \\ \hline
    First layer & Won & Drawn, Lost \\ 
    Second layer & Won, Drawn & Lost \\ 
    \bottomrule
    \end{tabular}
\end{center}
\end{table}

\subsection{Second-Layer Enhancements}

We have to decide how each of the previously proposed enhancements can benefit from the additional PN layer. The simplest case is for the solver part. The only difference is that instead of marking as resolved nodes that are proven on the first layer, we skip them if they are proven on the second layer, as it is easier to prove the superset. The remaining improvements are described in their respective sections and accompanied by the experiments testing their effectiveness.

All the experiments regarding double-layer PN-MCTS are performed on Awari, as it is the only game from our test domain with a significant number of draws.

\subsection{UCT-PN Rank Sorting}

For the UCT-PN part, there are multiple methods to leverage this new knowledge. One that seems straightforward is to enhance Formula~(\ref{UCT-PN_formula}), and introduce a new parameter $C_{pn2}$ that will rank nodes and influence the equation independently of the first layer. There are two downsides to this approach. One is that it introduces a new parameter that requires additional tuning. Second, in our opinion worse, is that it requires separate sorting of nodes to compute ranks. This will heavily influence the computational efficiency, which is already a~problem. 

Thus, our proposed solution is to modify the sorting mechanism, and use second-layer (dis) proof numbers as a tiebreaker. To ensure that this idea works, we took all variants that use UCT-PN formula and compare them with the corresponding results of the single-layer PN-MCTS from Table~\ref{tab:variants}. All variants with the final move selection use algorithm described before, taking into account only first-layer proven nodes. The results of this experiment are presented in Table~\ref{tab:2layer1}.

\begin{table}[h!!]
\begin{center}
    \caption{Influence of introducing second layer tie-breaker in the UCT-PN formula. Single-Layer/Double-Layer PN-MCTS  against a basic MCTS (Awari, 1000 games, $C_{pn}=1$, 1s per turn).}
    \label{tab:2layer1}
    \begin{tabular}{l|rr}
    \toprule
     \multirow{2}{*}{Variant}& \multicolumn{2}{c}{UCT-PN formula} \\
     & Single-layer & Double-layer  \\ \hline
    \texttt{xxU} & $48.2\pm2.69$ & $48.4\pm2.77$ \\ 
    \texttt{xSU} & $40.0\pm3.01$ & $47.1\pm2.85$ \\ 
    \texttt{FxU} & $49.8\pm2.72$ & $48.6\pm2.70$  \\ 
    \texttt{FSU} & $40.6\pm3.01$ & $46.4\pm2.83$ \\ 
    \bottomrule
    \end{tabular}
\end{center}
\end{table}

As we can see, the results are strictly better on the variants where this new sorting method is paired with a solver. In the remaining cases, the confidence bounds of both algorithms' winrates are overlapping, so they are tied. 
Thus, although the proposed second-layer UCT-PN is not a general improvement, it does not seem to be harmful, and it may lead to better results for some variants.

\subsection{Final Move Selection Contempt Factor}

Additional knowledge provided by second-layer proof numbers allows us to choose a move that leads to a proven draw if one is available. However, it is not a straightforward decision when to do so, as by choosing such a move, we usually give up any chances to win a game.

Thus, the so-called \emph{Contempt Factor} is used to determine when to prefer draw over the other lines of play. In our case, if the root score is strictly smaller than a contempt factor, then the proven draw child (if available) is selected. 

\begin{figure}[t]
    \centering
    \includegraphics[width= \columnwidth]{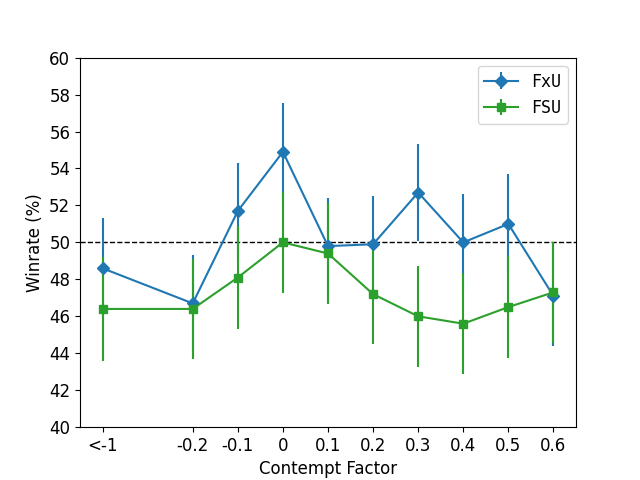}
    \caption{Comparison of various contempt factor values for Double-Layer PN-MCTS (Awari, 1000 games, $C_{pn}=1$, 1s per turn).}
    \label{fig:contempt}
\end{figure}

To test the impact of this parameter, we chose two variants which use final move selection (\texttt{FxU}, \texttt{FSU}), and extend the selection by different values of contempt factor: $<$-1, -0.2, 0, 0.2, 0.4, and 0.6. Note that the possible outcomes of the game are -1, 0, and 1. Thus, the contempt factor less than -1 behaves exactly as a single-layer PN-MCTS final move selection. The results of this experiment are presented in Figure~\ref{fig:contempt}.

As we can see, the contempt factor has a non-negligible impact on the algorithm's performance. The right value, which in the case of two of the tested variants would be around 0.0, increases the scores by 3.6--6.3\% compared to the standard single-layer PN final move selection, allowing \texttt{FxU} a confident win against the standard MCTS.

Thus, both presented improvements obtained by extending PN-MCTS by another layer of Proof-Number Search seem to positively impact the algorithm behavior---increasing its winrate given the right conditions and parameter values.

\section{Conclusions and Future Research}\label{sec:Conc}

This paper introduces PN-MCTS, a new Monte-Carlo Tree Search enhancement that combines it with
Proof-Number Search. We proposed several variants of the algorithm 
by taking advantage of the additional knowledge in three areas: during the final move selection, during the MCTS selection phase, and in the UCB1 formula. In particular, the ranking of the nodes according to their (dis)proof numbers is used to bias the UCB1 formula.

While there is a computational cost necessary to obtain the proof and disproof numbers, for the right domain and parameter values, the benefits can outweigh the costs. 
The results show that for LOA, MiniShogi, and Knightthrough, one can choose a variant of PN-MCTS that outperforms standard UCT for all time settings tested, some of which reach a winrate up to 96\%. An interesting case is Awari, where many tested variants are losing or drawing, but, especially for time limits below 1 second per move, PN-MCTS can convincingly win.

Because PNS can only deal with binary outcomes, in the basic PN-MCTS implementation, the disproof number is the same for a draw and a loss, potentially steering the search in the wrong direction. We aimed to resolve this issue by introducing a second layer of proof numbers to distinguish draws from losses. With this enhancement, double-layer PN-MCTS was able to win against standard MCTS in Awari for the same setting that was previously a tie.

There are multiple directions for future research. One is to test PN-MCTS on more domains and to investigate the reason why it does (not) work for certain games. Concurrently, one could test more parameter values with different time settings, and develop alternative methods of combining UCB1 formula with PN values. Although the ranking idea worked best from what we have tested so far, further studies on the subject are required. 
Eventually, we could search for more elaborate enhancements, taking advantage of the concept of multi-layer PN-MCTS. In particular, extend the double-layer variant into a multi-layer one, to handle games with multiple outcomes. Alternatively, this concept can be adapted to handle games with more than two players by computing proven subtrees against each of them separately.

We also plan to extend our research to Product Propagation \cite{Slagle1968Experiments}. Another approach to backup probabilistic information in two-player game tree search, shown to outperform PNS on some games \cite{saffidine2014developments}. 
Thus, it is a natural line of research to test if and how our PNS-related developments will transfer to Product Propagation.

\section{Acknowledgments}

This research is partially funded by the European Research Council as part of the Digital Ludeme Project (ERC Consolidator Grant \#771292).

This research was also supported in part by the National Science Centre, Poland, under project number 2021/41/B/ST6/03691 (Jakub Kowalski). 
For the purpose of Open Access, the author has applied a CC-BY public copyright licence to any
Author Accepted Manuscript (AAM) version arising from this submission.

\bibliographystyle{IEEEtran} 
\bibliography{refs,references} 


\onecolumn


\appendix

\vspace{2cm}

\begin{table*}[h!]
\centering
\caption{Exact winrate values for the $C_{pn}$ tuning experiment  shown in Figure~\ref{fig:cpntuning}.}
\label{tab:cpntuning}
\begin{tabular}{c|c|c|c|c|c|c|c}
\toprule
\multirow{2}{*}{Game}& \multicolumn{7}{c}{$C_{pn}$ value} \\ 
 & 0.0 & 0.1 & 0.5 & 1.0 & 2.0 & 5.0 & $10^6$ \\ \hline 
LOA 7$\times$7  & $46.0\pm3.09$ & $64.1\pm2.97$ & $86.9\pm2.09$ & $90.4\pm1.83$ & $91.6\pm1.72$ & $82.2\pm2.37$ & $69.0\pm2.87$ \\
LOA 8$\times$8  & $45.2\pm3.09$ & $55.9\pm3.08$ & $81.9\pm2.39$ & $83.2\pm2.32$ & $80.5\pm2.46$ & $57.4\pm3.06$ & $32.3\pm2.88$ \\
\bottomrule
\end{tabular}

\vspace{2cm}

  \centering
\caption{Exact winrate values for the different time settings comparison shown in Figure~\ref{fig:timetrends}.}
\label{tab:timetrends}
\begin{tabular}{c||c|c|c|c|c|c|}
\toprule
\multicolumn{7}{c}{LOA $7\times 7$} \\ \hline
Variant & $1/8s$ & $1/4s$ & $1/2s$ & $1s$ & $2s$ & $4s$ \\ 
\verb|FSx| & $63.3\pm2.99$ & $\textbf{70.9}\pm2.82$ & $67.2\pm2.91$ & $65.6\pm2.95$ & $58.4\pm3.06$ & $53.5\pm3.09$ \\
\verb|FSU| & $85.4\pm2.19$ & $90.4\pm1.83$ & $91.9\pm1.69$ & $\textbf{93.2}\pm1.56$ & $91.0\pm1.78$ & $84.0\pm2.27$ \\
\verb|FxU| & $72.0\pm2.78$ & $77.3\pm2.60$ & $83.5\pm2.30$ & $90.4\pm1.83$ & $\textbf{90.6}\pm1.80$ & $83.6\pm2.30$ \\ 
\verb|xSU| & $67.0\pm2.92$ & $84.1\pm2.27$ & $92.8\pm1.60$ & $91.1\pm1.76$ & $91.5\pm1.73$ & $87.2\pm2.07$ \\ \hline
\multicolumn{7}{c}{LOA $8\times 8$} \\ \hline
Variant & $1/8s$ & $1/4s$ & $1/2s$ & $1s$ & $2s$ & $4s$ \\ 
\verb|FSx| & $51.9\pm3.10$ & $57.5\pm3.07$ & $65.6\pm2.95$ & $\textbf{66.4}\pm2.93$ & $59.0\pm3.05$ & $53.4\pm3.09$\\
\verb|FSU| & $60.0\pm3.04$ & $72.7\pm2.76$ & $85.4\pm2.19$ & $90.8\pm1.79$ & $\textbf{96.2}\pm1.19$ & $95.3\pm1.31$\\
\verb|FxU| & $59.2\pm3.05$ & $58.0\pm3.06$ & $73.4\pm2.74$ & $83.2\pm2.32$ & $91.5\pm1.73$ & $\textbf{94.4}\pm1.43$\\ 
\verb|xSU| & $55.4\pm2.87$ & $63.3\pm2.99$ & $82.7\pm2.35$ & $\textbf{92.9}\pm1.59$ & $93.7\pm1.51$ & $94.5\pm1.40$ \\ \hline
\multicolumn{7}{c}{MiniShogi} \\ \hline
Variant & $1/8s$ & $1/4s$ & $1/2s$ & $1s$ & $2s$  & $4s$ \\ 
\verb|FSx| & $57.4\pm3.07$ & $53.9\pm3.09$ & $52.4\pm3.10$ & $53.2\pm3.09$ & $\textbf{60.3}\pm3.03$ & $56.1\pm3.08$ \\
\verb|FSU| & $58.1\pm3.06$ & $57.4\pm3.07$ & $63.0\pm2.99$ & $\textbf{65.8}\pm2.94$ & $64.5\pm2.97$ & $58.6\pm3.05$ \\
\verb|FxU| & $57.8\pm3.06$ & $56.8\pm3.07$ & $59.7\pm3.04$ & $\textbf{64.5}\pm2.97$ & $59.1\pm3.05$ & $54.1\pm3.09$ \\ 
\verb|xSU| & $18.6\pm2.41$ & $20.5\pm2.50$ & $38.6\pm3.02$ & $60.3\pm3.03$ & $59.8\pm3.04$ & $56.6\pm3.07$\\ \hline
\multicolumn{7}{c}{Knightthrough} \\ \hline
Variant & $1/8s$ & $1/4s$ & $1/2s$ & $1s$ & $2s$  & $4s$ \\ 
\verb|FSx| & $\textbf{79.9}\pm2.49$ & $73.0\pm2.75$ & $58.6\pm3.05$ & $59.2\pm3.05$ & $68.4\pm2.88$ & $78.1\pm2.56$ \\
\verb|FSU| & $\textbf{83.8}\pm2.28$ & $77.7\pm2.58$ & $62.9\pm3.00$ & $66.8\pm2.92$ & $68.2\pm2.89$ & $72.3\pm2.78$ \\
\verb|FxU| & $50.4\pm3.10$ & $54.1\pm3.09$ & $\textbf{56.9}\pm3.07$ & $56.4\pm3.08$ & $50.5\pm3.10$ & $50.6\pm3.10$ \\ 
\verb|xSU| & $62.0\pm3.01$ & $83.3\pm2.31$ & $66.8\pm2.92$ & $63.5\pm2.99$ & $66.2\pm2.93$ & $72.0\pm2.78$\\ \hline
\multicolumn{7}{c}{Awari} \\ \hline
Variant & $1/8s$ & $1/4s$ & $1/2s$ & $1s$ & $2s$  & $4s$ \\ 
\verb|FSx| & $\textbf{39.9}\pm3.01$ & $39.4\pm2.99$ & $38.1\pm2.93$ & $38.0\pm2.94$ & $32.7\pm2.84$ & $28.2\pm2.71$\\
\verb|FSU| & $\textbf{56.6}\pm3.02$ & $51.9\pm3.07$ & $41.6\pm3.02$ & $40.6\pm3.01$ & $33.6\pm2.90$ & $25.1\pm2.67$\\
\verb|FxU| & $\textbf{64.8}\pm2.78$ & $57.6\pm2.84$ & $55.0\pm2.71$ & $49.8\pm2.72$ & $48.0\pm2.59$ & $40.8\pm2.36$\\ 
\verb|xSU| & $65.9\pm2.89$ & $55.5\pm3.05$ & $45.1\pm3.06$ & $40.0\pm3.01$ & $37.1\pm2.98$ & $31.9\pm2.87$ \\
\bottomrule
\end{tabular}

\vspace{2cm}

\caption{Exact winrate values for the different contempt factor comparison shown in Figure~\ref{fig:contempt}.}
\label{tab:contempt}
\hspace{-45px}\begin{tabular}{c|c|c|c|c|c|c|c|c|c|c}
\toprule
\multirow{2}{*}{Variant}& \multicolumn{10}{c}{Contempt Factor} \\
 & $<$-1.0 & -0.2 & -0.1 & 0.0 & 0.1 & 0.2 & 0.3 & 0.4 & 0.5 & 0.6\\ \hline 
\verb|FxU|  & $48.6\pm2.70$ &  $46.7\pm2.62$ & $51.7\pm2.62$ & $\textbf{54.9}\pm2.66$ & $49.8\pm2.60$ & $49.9\pm2.63$ & $52.7\pm2.65$ & $50.0\pm2.61$ & $51.0\pm2.68$ & $47.1\pm2.68$ \\
\verb|FSU|  & $46.4\pm2.83$ & $46.4\pm2.71$ & $48.1\pm2.78$ & $\textbf{50.0}\pm2.74$ & $49.4\pm2.75$ & $47.2\pm2.72$ & $46.0\pm2.73$ & $45.6\pm2.74$ & $46.5\pm2.77$ & $47.3\pm2.72$ \\
\verb|xSU|  & $47.1\pm2.85$ & $46.4\pm2.85$  & $47.4\pm2.85$ & $46.7\pm2.86$ & $45.8\pm2.81$ & $46.4\pm2.84$ & $47.4\pm2.82$ & $\textbf{48.0}\pm2.86$ & $44.6\pm2.82$  & $44.9\pm2.84$  \\
\bottomrule
\end{tabular}
\end{table*}

\end{document}